\documentclass[10pt,twocolumn,letterpaper]{article}

\usepackage{cvpr}              

\usepackage[dvipsnames]{xcolor}

\usepackage{dsfont}
\newcommand{\pub}[1]{\color{black}{\scriptsize{{#1}}}}
\usepackage{multicol}
\usepackage{multirow}
\usepackage{stfloats}
\usepackage[accsupp]{axessibility}

\definecolor{cvprblue}{rgb}{0.21,0.49,0.74}
\usepackage[pagebackref,breaklinks,colorlinks,citecolor=cvprblue]{hyperref}

\def\paperID{11782} 
\def\confName{CVPR}
\def\confYear{2024}

\title{Constructing and Exploring Intermediate Domains in Mixed Domain Semi-supervised Medical Image Segmentation}

\author{
Qinghe Ma$^1$,~~Jian Zhang$^1$,~~Lei Qi$^2$,~~Qian Yu$^3$,~~Yinghuan Shi$^{1,}$\thanks{Corresponding author. 
Qinghe Ma, Jian Zhang, Yinghuan Shi and Yang Gao are with the State Key Laboratory for Novel Software Technology and National Institute of Healthcare Data Science, Nanjing University, China. 
This work was supported by the NSFC Program (62222604, 62206052, 62192783), Jiangsu Natural Science Foundation (BK20210224), and Shandong Natural Science Foundation (ZR2023MF037).
}~,~~Yang Gao$^1$\\
$^1$Nanjing University~~~~~$^2$Southeast University~~~~~$^3$Shandong Women's University\\
}

\begin{document}
\maketitle
\begin{abstract}
Both limited annotation and domain shift are prevalent challenges in medical image segmentation. Traditional semi-supervised segmentation and unsupervised domain adaptation methods address one of these issues separately. However, the coexistence of limited annotation and domain shift is quite common, which motivates us to introduce a novel and challenging scenario: \textbf{Mi}xed \textbf{D}omain \textbf{S}emi-supervised medical image \textbf{S}egmentation (MiDSS). In this scenario, we handle data from multiple medical centers, with limited annotations available for a single domain and a large amount of unlabeled data from multiple domains. We found that the key to solving the problem lies in how to generate reliable pseudo labels for the unlabeled data in the presence of domain shift with labeled data. To tackle this issue, we employ Unified Copy-Paste (UCP) between images to construct intermediate domains, facilitating the knowledge transfer from the domain of labeled data to the domains of unlabeled data. To fully utilize the information within the intermediate domain, we propose a symmetric Guidance training strategy (SymGD), which additionally offers direct guidance to unlabeled data by merging pseudo labels from intermediate samples. Subsequently, we introduce a Training Process aware Random Amplitude MixUp (TP-RAM) to progressively incorporate style-transition components into intermediate samples. Compared with existing state-of-the-art approaches, our method achieves a notable 13.57\% improvement in Dice score on Prostate dataset, as demonstrated on three public datasets. Our code is available at \textcolor{magenta}{\url{https://github.com/MQinghe/MiDSS}}
\end{abstract}
\section{Introduction}
\begin{figure}[!t]
\centering
\includegraphics[width=1.0\columnwidth]{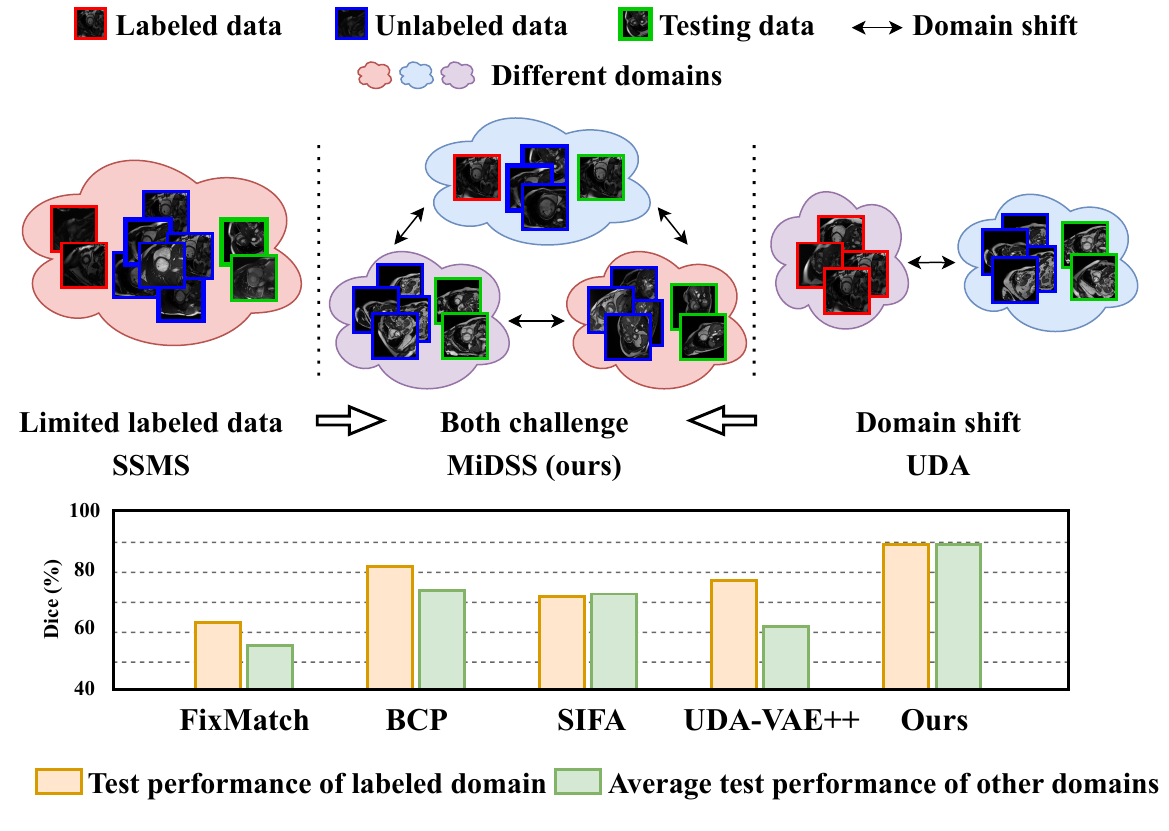}
\caption{The upper figure illustrates SSMS, UDA, and MiDSS. 
The lower figure shows the comparison between different methods on the labeled domain (BIDMC)~\cite{liu2020shape} and other domains. 
}
\label{setting}
\vspace{-10pt}
\end{figure}
Semi-supervised medical image segmentation (SSMS) has aroused lots of attention in recent years, due to its advantage of improving performance while reducing the labeling burden~\cite{fan2020inf, li2020transformation, yu2019uncertainty, li2022pln}. Surprisingly, the performance of semi-supervised methods~\cite{li2023iomatch,duan2023towards} is sometimes close to or even better than that of its upper bound---supervised methods~\cite{shi2020clinically,chen2021deep}. With 1) only a few labeled data as guidance, and 2) the development of unsupervised learning techniques, \eg, contrastive learning~\cite{he2020momentum}, SSMS is becoming a promising direction to utilize unlabeled data. In traditional SSMS tasks, it is typically assumed that labeled and unlabeled data come from the same distribution~\cite{bai2017semi,you2022simcvd,cai2023orthogonal}. In this case, the model applies the information gained from labeled data directly to predict pseudo labels for unlabeled data.



Despite their success, we notice a significant yet easily underestimated issue: \emph{During clinical data collection, do we really check and guarantee, whether each unlabeled sample belongs to the same distribution as labeled samples?} For example, given a certain amount of labeled data collected and well annotated from a typical center, since unlabeled data are easily accessed and collected, when the physicians have a massive number of unlabeled data, they possibly lack sufficient time to verify whether the data belongs to the same distribution~\cite{dash2019big, rudrapatna2020opportunities}. 


As this dilemma mentioned above, which refers to ``domain shift"~\cite{wang2019semi,guan2021domain}, could cause a large performance degeneration, we should pay more attention to this issue during our development of a general SSMS model. A more practical setting could be: given labeled data from a consistent distribution, the unlabeled data is allowed to come from the same distribution or even multiple different distributions. 
It is important to note that this issue could not be directly solved by current unsupervised domain adaptation (UDA) methods. Although UDA is promising to address domain shift, these methods~\cite{chen2020unsupervised,russo2018source,dou2018unsupervised,yang2020fda} demand abundant labeled source domain data, which is the primary reason for the existence of SSMS. 

Despite the significance of the domain shift issue in SSMS, according to our best knowledge, our work is first to attempt to investigate this issue, which we termed as \textbf{Mi}xed \textbf{D}mains \textbf{S}emi-supervised medical image \textbf{S}egmentation (abbreviated as \textbf{MiDSS} in our paper). As illustrated in \cref{setting}, confronted with the coexistence of limited annotation and domain shift, it is challenging to effectively utilize a large quantity of unlabeled medical images from multiple domains. Additionally, from the lower figure in \cref{setting}, we found that achieving satisfactory results in our setting is difficult when relying solely on SSMS or UDA methods.


The key to addressing the problem lies in \textit{how to generate reliable pseudo labels for unlabeled data, especially in the presence of domain shift}. However, the domain gap among training samples results in the model producing poor pseudo labels for unlabeled data.
Thus, our key insight is first to construct several intermediate domains among original domains to narrow the domain gap, and then emphasize knowledge transfer approaches upon these constructed domains.
CutMix~\cite{yun2019cutmix}, also referred to as Copy-Paste (CP), is a simple yet highly effective technique for generating intermediate samples by embedding patches from labeled samples into unlabeled samples along a single direction.
To fully leverage the data generation capability of CP, we employ a Unified Copy-Paste (UCP) approach. UCP generates a great quantity of diverse intermediate samples by unified applying both embedding directions between labeled and unlabeled samples, narrowing the domain gap and effectively mitigating the issue of error accumulation. 

Besides, the intermediate domains may not align perfectly with the domains of the unlabeled data.
Solely training within the intermediate domains neglects the model performance on the unlabeled data, which are of great interest. 
The information from intermediate domains should be harnessed to promote the model to generate high quality pseudo labels for unlabeled data. 
Moreover, medical images of the same organ or lesion often exhibit similar structures~\cite{wang2022separated}, with stylistic differences being the primary source of domain shift~\cite{liu2021feddg}. Local semantic mixing methods (\eg, CP) fail to address these differences, leading to a lack of stylistic transition in intermediate domains. Additionally, it is crucial that an aggressive stylistic transition hinders the gradual construction of intermediate domains.

Therefore, to fully harness the information from intermediate domains and ensure comprehensive and stable knowledge transfer, we propose a Mixed Domains Semi-supervised medical image segmentation approach. In terms of training strategy, we design Symmetric Guidance training strategy (SymGD), which is composed of guidance from unlabeled data to intermediate samples and vice versa. This dual-perspective integration of pseudo labels enhances the precision of guidance. Concerning intermediate samples, we introduce a Training Process aware Random Amplitude MixUp module (TP-RAM) to promote smooth stylistic knowledge transfer. Our main contributions are three folds:
\begin{itemize}
    \item We investigate a new yet underestimated semi-supervised medical image segmentation setting (namely MiDSS).
    \item A novel SymGD training strategy based on UCP, promotes the training on the unlabeled data with intermediate domains information.
    \item A TP-RAM module to make domain knowledge transfer comprehensive and stable.
\end{itemize}

Extensive experiments validate the effectiveness of our method on three public datasets.
For example, on Prostate dataset, we achieved a remarkable Dice score improvement of 13.57\% compared with all state-of-the-art methods.

\section{Related Work}

\begin{figure*}[!t]
\centering
\includegraphics[width=0.9\linewidth]{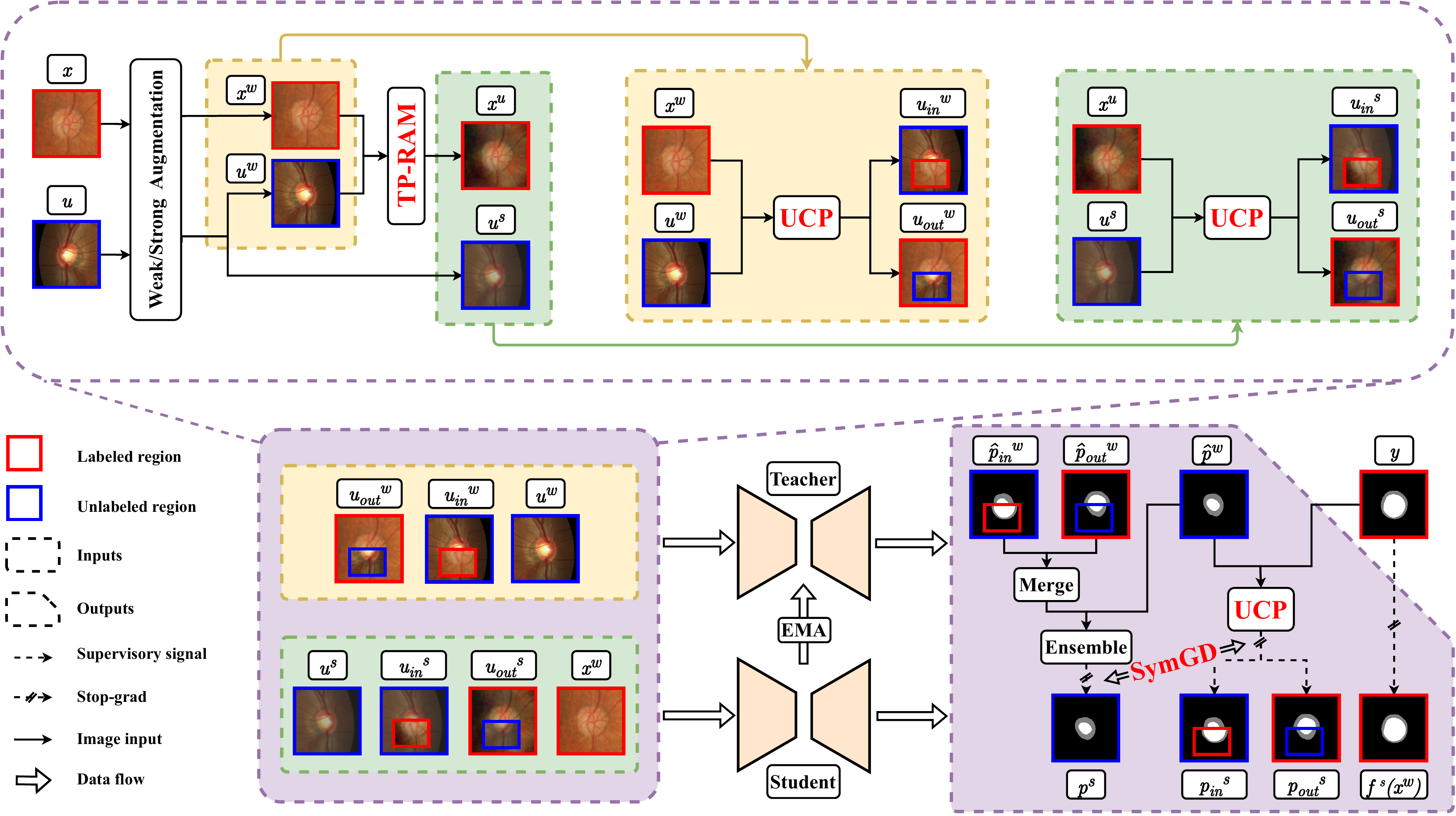}
\caption{The overall framework of our method emphasizes domain knowledge transfer through data augmentation and training strategy. We generate intermediate samples through UCP between labeled data and unlabeled data. During training, we gradually introduce style transfer components to the intermediate samples, constructing intermediate domains at both semantic and stylistic levels. Further details about UCP and TP-RAM are provided in \cref{ucp} and \cref{tpram}. Moreover, we design a symmetric guidance for model training. In addition to guiding from unlabeled data to intermediate samples, we merge unlabeled regions of intermediate samples to obtain the pseudo label of unlabeled data from another perspective. The integration of pseudo labels through ensemble from two perspectives guides the prediction of unlabeled data. Best viewed in color.}
\label{framework}
\vspace{-5pt}
\end{figure*}

\textbf{Semi-supervised Medical Image Segmentation.} Manually annotating medical images is challenging and costly~\cite{zhuang2013challenges}. Semi-supervised medical image segmentation methods have shown promise in addressing limited-label segmentation tasks. Entropy minimization and consistency regularization are widely adopted techniques in this context. 
Yu \etal~\cite{yu2019uncertainty} encouraged consistent predictions under different perturbations. Luo \etal~\cite{luo2021semi} proposed a dual-task network that predicts pixel-wise segmentation maps and geometry-aware level set representations. Li \etal~\cite{li2020shape} introduced a shape-aware strategy using a multi-task deep network that jointly predicts semantic segmentation and signed distance maps. 
Wu \etal~\cite{wu2022exploring} addressed challenges in SSMS by simultaneously enforcing pixel-level smoothness and inter-class separation.
Miao \etal~\cite{miao2023caussl} pointed out the importance of algorithmic independence between two networks or branches in SSMS. Unfortunately, existing methods often face challenges when domain shift occurs, causing decreased performance due to the shared distribution assumption between labeled and unlabeled data.

\textbf{Unsupervised Domain Adaptation.} UDA methods train the model by leveraging abundant labeled source domain data and unlabeled target domain data to achieve appealing performance in the target domain. In medical image segmentation tasks, domain shift issues arise from variations in device parameters, disease severity, imaging principles, \etc. Thus, UDA plays a crucial role in addressing these issues. Adversarial learning based UDA
methods achieve alignment at multiple levels to narrow domain gap, including input alignment \cite{russo2018source,chartsias2017adversarial} and feature alignment \cite{ganin2016domain,dou2018unsupervised,zhao2022uda,tsai2018learning}. Self-training based UDA methods \cite{zheng2021rectifying, zhang2020collaborative} train the model by generating reliable pseudo labels for unlabeled data. Methods like \cite{chen2022deliberated,yang2020fda} promote domain knowledge transfer by generating intermediate domains. These methods typically rely on a sufficient amount of source domain data and focus on a single target domain, making it challenging to be effective in the MiDSS scenario.

\textbf{Data Augmentation via Copy-Paste.} Copy-paste (CP)~\cite{yun2019cutmix} involves pasting the content of a particular region of one image onto the corresponding region of another image, creating a new image that retains the semantic information from both original images. Compared to other pixel-level fusion strategies, such as MixUp~\cite{zhang2017mixup}, CP excels in preserving and blending semantic information from source images. In contrast, MixUp combines global source images proportionally, potentially leading to ambiguity issues when pixels from different classes are blended. Therefore, CP is a more suitable augmentation technique for medical image segmentation tasks. In line with this research direction, BCP~\cite{bai2023bidirectional} takes into account the dual embedding directions between unlabeled and labeled samples, randomly selecting one of them to generate intermediate samples. 
To mitigate domain shift, previous works~\cite{fan2022ucc,bai2023bidirectional} employed CP to transfer knowledge from labeled data to unlabeled data by generating intermediate samples. Despite the progress, they fall short in fully leveraging information from intermediate domains, leading to sub-optimal transfer effects.
\vspace{-5pt}
\section{Method}
In the MiDSS scenario, the domain gap exists among images originating from $K$ data centers $\{\mathcal{D}_i\}_{i=1}^K$. The training set comprises $N$ labeled images $\{(x_i,y_i)\}_{i=1}^N$ from a single domain $\mathcal{D}_j \in \{\mathcal{D}_i\}_{i=1}^K$ and $M$ unlabeled images $\{u_i\}_{i=1}^M$ from multiple domains $\mathcal{D}_1,\ldots,\mathcal{D}_K$, where $M > N$. The image resolution of $H \times W \times D$ sequentially represents height and width and channel, and $y_i \in \{0,1\}^{H \times W \times C}$ is the ground truth of $x_i$, where $C$ is the number of class. We aim to train a model based on the aforementioned training set, capable of delivering outstanding segmentation performance across all domains with or without labeled data. 


Based on the mean teacher framework~\cite{tarvainen2017mean}, for each labeled data $x$, 
teacher model $f^t$ generates probability map $p^w$ and pseudo label $\hat{p}$ for weakly augmented unlabeled data $u^w$, guiding the student model $f^s$ in making predictions $p^s$ for strongly augmented unlabeled data $u^s$: 
\begin{equation}
    \begin{split}
        &x^w=\mathcal{A}^w(x); u^w=\mathcal{A}^w(u); u^s=\mathcal{A}^s(u),\\
        &p^w =f^t(u^w); p^s=f^s(u^s); \hat{p}=\arg\max(p^w),
    \end{split}
\end{equation}
where $\arg\max(\cdot)$ applied to probability distributions produces valid ``one-hot" probability distributions, and $f^t$ is updated by the EMA of $f^s$~\cite{tarvainen2017mean}. The weak augmentation $\mathcal{A}^w$ includes cropping, rotation, flip, and elastic distortion, while the strong augmentation $\mathcal{A}^s$ builds upon $\mathcal{A}^w$ by incorporating non-geometric operations like color jitter and Gaussian blurring~\cite{sohn2020fixmatch,upretee2022fixmatchseg}. The overall framework of our method is shown in~\cref{framework}.


\subsection{Intermediate Samples Generation by UCP}


\begin{figure}[!t]
\centering
\includegraphics[width=0.9\columnwidth]{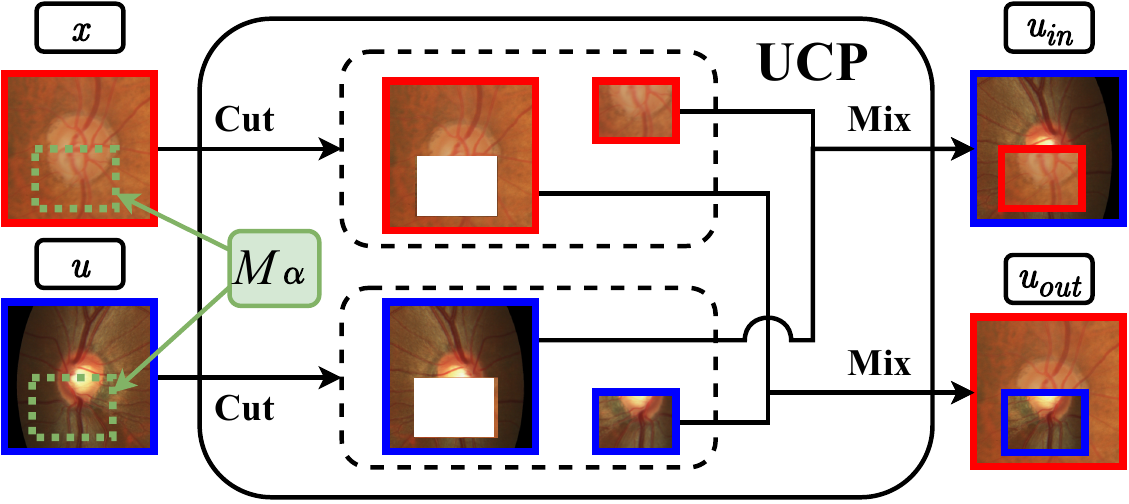}
\caption{The illustration of UCP between images. \textbf{Cut} refers to splitting the image into two parts according to $M_{\alpha}$, while \textbf{Mix} implies merging two parts of images back together.}
\label{ucp}
\end{figure}


Given a pair of labeled data $(x^w,y^w)$ and unlabeled data $(u^s,p^w,\hat{p})$, we simultaneously apply both embedding directions of CP. This enables unified Copy-Paste (UCP) between labeled and unlabeled data with distribution discrepancy. Through this approach, we obtain intermediate samples along with their pseudo-labels and probability maps:


\begin{equation}
    \label{CPformula}
    \begin{gathered}
        u^s_{in}=x^w \odot M_\alpha + u^s \odot (\mathbf{1}-M_\alpha),\\
        p^w_{in}=y^w \odot M_\alpha + p^w \odot (\mathbf{1}-M_\alpha),\\
        \hat{p}_{in}=y^w \odot M_\alpha + \hat{p} \odot (\mathbf{1}-M_\alpha),\\
        u^s_{out}=u^s \odot M_\alpha + x^w \odot (\mathbf{1}-M_\alpha),\\
        p^w_{out}=p^w \odot M_\alpha + y^w \odot (\mathbf{1}-M_\alpha),\\
        \hat{p}_{out}=\hat{p} \odot M_\alpha + y^w \odot (\mathbf{1}-M_\alpha),
    \end{gathered}
\end{equation}
where $M_\alpha\in\{0,1\}^{W \times H}$ is a randomly generated one-centered mask, indicating the region for CP operation. In this context, $\mathbf{1}$ indicates an all-one matrix, and $\odot$ means element-wise multiplication. Referring to \cref{ucp}, the illustration explains UCP between images, with the same operation applied to both probability maps and pseudo labels.



\subsection{Symmetric Guided Training with UCP}
In the case of a labeled data $(x^w, y^w)$ and unlabeled data $(u^w, u^s)$ pair, we aim to facilitate the transfer of knowledge from the labeled domain while maximizing the extraction of latent information from the unlabeled data. 

During the training phase, the teacher model firstly predicts $u^w$ to generate probability map $p^w$ and pseudo label $\hat{p}$ for $u^s$. Then we employ UCP between $(x^w, y^w)$ and $(u^s, p^w, \hat{p})$ to generate intermediate samples $(u^{s}_{in},u^{s}_{out})$, probability maps $(p^{w}_{in},p^{w}_{out})$ and pseudo labels $(\hat{p}_{in},\hat{p}_{out})$ by \cref{CPformula}. We set weight map $w$ to indicate whether the pseudo label is reliable:
\begin{equation}
    \label{pseudo_label_weighted}
    w_i=\mathds{1}(\max (p^w_i) \ge \tau),
\end{equation}
where $w_i$ is $i^{th}$ pixel of $w$, and $\tau$ is a pre-defined confidence threshold used to filter noisy labels. The indicator function is denoted as $\mathds{1}(\cdot)$. The pseudo labels $\hat{p}_{in}$ and $\hat{p}_{out}$ will be used as the supervision to guide the student model in predicting $p^{s}_{in}$ and $p^{s}_{out}$ for $u^{s}_{in}$ and $u^{s}_{out}$:
\begin{equation}
    \begin{gathered}
        \mathcal{L}_{in}=\mathcal{L}_{ce}(\hat{p}_{in}, p^{s}_{in},w_{in})+\mathcal{L}_{dice}(\hat{p}_{in}, p^{s}_{in}, w_{in}),\\
        \mathcal{L}_{out}=\mathcal{L}_{ce}(\hat{p}_{out}, p^{s}_{out}, w_{out})+\mathcal{L}_{dice}(\hat{p}_{out}, p^{s}_{out}, w_{out}),
    \end{gathered}
\end{equation}
where $\mathcal{L}_{ce}$ and $\mathcal{L}_{dice}$ respectively represent the cross-entropy loss and dice loss, which are formulated as:
\begin{equation}
    \begin{gathered}
        \mathcal{L}_{ce}(y,p,w)=-\frac{1}{H \times W}\sum_{i=1}^{H \times W} w_i y_i\log p_i,\\
        \mathcal{L}_{dice}(y,p,w)=1-\frac{2\times \sum_{i=1}^{H \times W}w_i p_i y_i}{\sum_{i=1}^{H \times W} w_i(p_i^2+y_i^2)},
    \end{gathered}
\end{equation}
where $p_i$, $y_i$ denote the probability of foreground and pseudo label of the $i^{th}$ pixel, respectively.

\begin{figure}[!t]
\centering
\includegraphics[width=0.9\columnwidth]{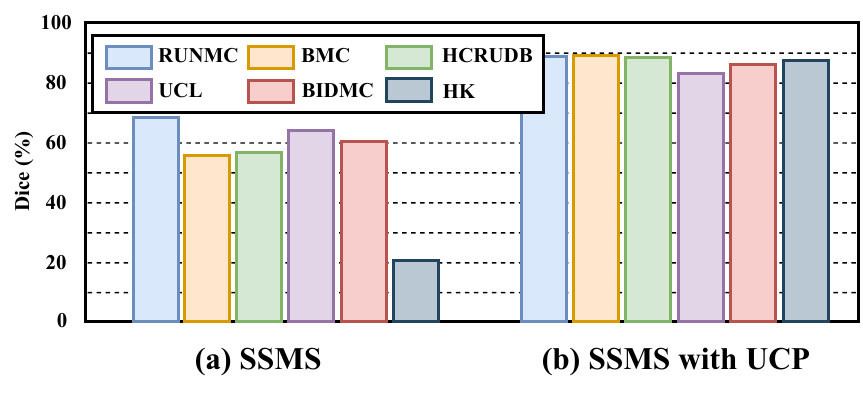}
\caption{The results depict the quality of pseudo labels in a SSMS method (FixMatch) with and without UCP, utilizing 40 labeled data from the BIDMC domain in Prostate dataset~\cite{liu2020shape}. Each colored bar represents samples from different domains, with the bar height indicating the quality of pseudo labels generated by the model for unlabeled data from that domain.}
\label{w_wo_ucp}
\vspace{-10pt}
\end{figure}

The supervision provided by the fusion of $\hat{p}$ with $y^w$ to $p^{s}_{in}$ and $p^{s}_{out}$ effectively promotes the model training in the intermediate domains, facilitating adaptation to the domains of unlabeled data. With the assistance of UCP, the quality of pseudo labels for unlabeled data in each domain is significantly improved as shown in \cref{w_wo_ucp}. However, this single-direction supervision lacks further utilization of information acquired in the intermediate domains, resulting in sub-optimal performance in the domains of unlabeled data. To address such limitation, we design a Symmetric Guidance training strategy (SymGD) to fully utilize the information from the intermediate domains. In addition to the consistency mentioned above, we also explore the direct guidance of unlabeled data through intermediate domain information.

Similarly, we employ $M_\alpha$ to mix $x^w$ and $u^w$, generating intermediate samples $u^{w}_{in}$ and $u^{w}_{out}$. The teacher model predicts them to generate pseudo labels $\hat{p}^{w}_{in}$ and $\hat{p}^{w}_{out}$. We merge the unlabeled regions of the pseudo labels:
\begin{equation}
    \label{pseudo_label_concat}
    \hat{p}^{mg} = \hat{p}^{w}_{out} \odot M_\alpha + \hat{p}^{w}_{in} \odot (\mathbf{1}-M_\alpha).
\end{equation}

We obtain the pseudo label for $u^s$ from the pseudo labels of intermediate samples, then integrate it with $p^w$ to provide more accurate guidance for the prediction of the student model on $u^s$. The weight map $w^{ens}$ of $\hat{p}^{mg}$ is determined as follows:
\begin{equation}
    \label{ensemble}
    w^{ens}=(1 - (\hat{p} \oplus \hat{p}^{mg})) \odot w \odot w^{mg},
\end{equation}
where $\oplus$ is the pixel-wise XOR operator to indicate the consistency between pseudo labels from two perspectives, and $w$ and $w^{mg}$ represent the weight map of $\hat{p}$ and $\hat{p}^{mg}$ respectively. The loss of such direction can be defined as:
\begin{equation}
    \label{symmetric loss}
    \mathcal{L}_{sym} = \mathcal{L}_{ce}(\hat{p}^{mg},p^s,w^{mg})+\mathcal{L}_{dice}(\hat{p}^{mg},p^s,w^{mg}),
\end{equation}

To generate reliable pseudo labels for unlabeled data from the perspective of intermediate samples, it is necessary that the model performs well in the intermediate domain. Therefore, the guidance from $\hat{p}^{mg}$ to $p^s$ maintains a relatively low weight in the early stages of training and rapidly increases in the middle to later stages.

\begin{figure}[!t]
\centering
\includegraphics[width=0.9\columnwidth]{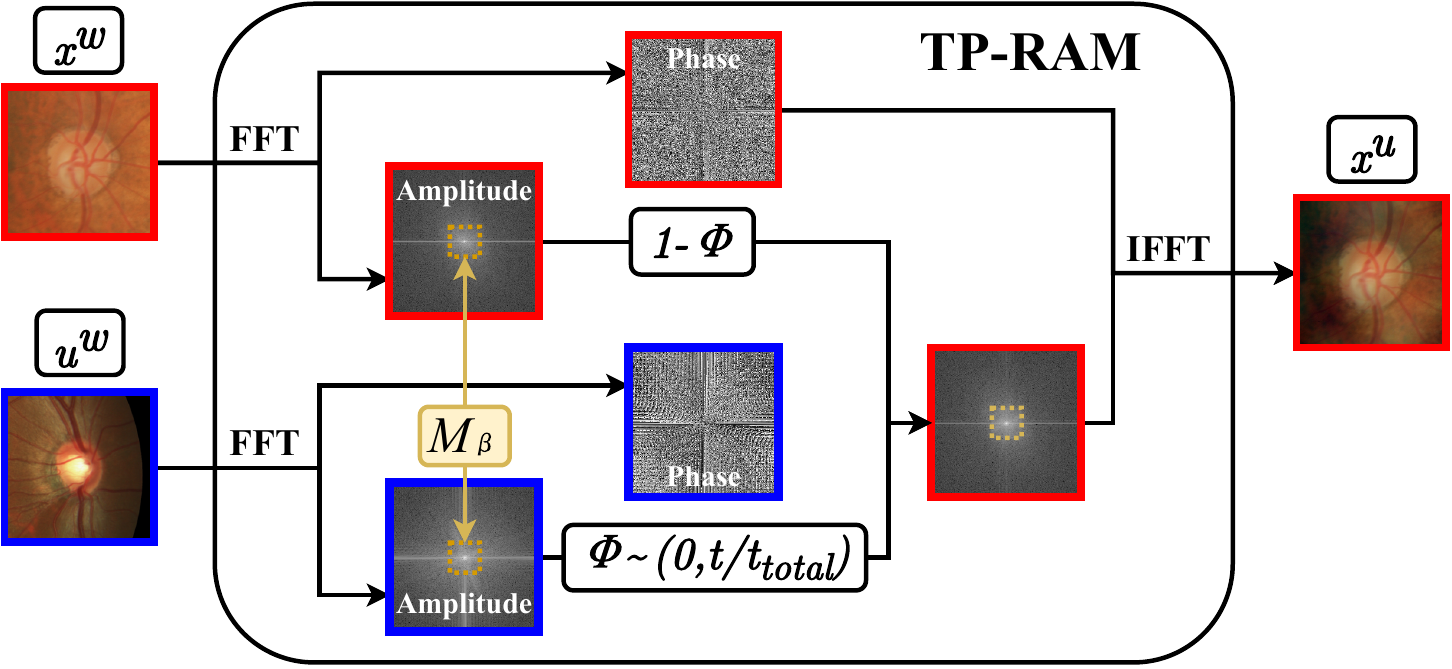}
\caption{The illustration of TP-RAM. Fast Fourier transform \textbf{(FFT)} extracts the amplitude and phase maps of $x^w$ and $y^w$. The low-frequency regions (determined by $M_\beta$) of the two phase maps are mixed. Through inverse Fast Fourier transform \textbf{(IFFT)}, $x^u$ is synthesized, preserving consistent semantics while introducing a different style compared to $x^w$.}
\label{tpram}
\vspace{-10pt}
\end{figure}

\subsection{Style Transition in Intermediate Samples}
The intermediate sample generation method based on UCP primarily focuses on local semantic blending while overlooking the transition of stylistic differences between different domains. Distribution information (\ie, style) is typically represented in low-frequency components, while edge information is often found in high-frequency components \cite{liu2021feddg}. 
By interpolating low-frequency information between data from different domains, amplitude MixUp generates samples with a new style. Then we mix these samples with $u_w$ by UCP to introduce style-transition components in the intermediate domains.

However, arbitrary style transfer is not conducive to the stable model learning of the intermediate domain. We aim for the intermediate samples to exhibit a gradual transition in style between domains during the training process without compromising the diversity introduced by random MixUp. Therefore, we propose Training Progress aware Random Amplitude Mixup (TP-RAM).

Overall, as exhibited in \cref{tpram}, we obtain the frequency space signals of each channel of the image $x$ through Fast Fourier transformation $\mathcal{F}$ as follows:
\begin{equation}
    \label{fourier}
    \mathcal{F}(x)(u,v)=\sum_{h=0}^{H-1}\sum_{w=0}^{W-1}x(h,w)e^{-j2 \pi (\frac{h}{H}u+\frac{w}{W}v)},j^2=-1.
\end{equation}
Let $\mathcal{F}^{A}$,$\mathcal{F}^{P}$ be the amplitude and phase components of $\mathcal{F}$. For a pair of labeled and unlabeled data pair $(x^w,u^w)$, we obtain their amplitude $(\mathcal{F}^A(x^w), \mathcal{F}^A(u^w))$ and phase $(\mathcal{F}^P(x^w), \mathcal{F}^P(u^w))$. After that, we blend the low-frequency information of $\mathcal{F}^A(u^w)$ into $\mathcal{F}^A(x^w)$, and obtain a new image $x^u$ through inverse Fast Fourier transformation $\mathcal{F}^{-1}$ as follows:
\begin{equation}
    \label{xu}
    x^u = \mathcal{F}^{-1}[M_\beta \odot \mathcal{F}^A(u^w) + (\mathbf{1}-M_\beta)\odot\mathcal{F}^A(x^w),\mathcal{F}^P(x^w)],
\end{equation}
where $M_{\beta}$ is a mask with values of $0$ except in the central $2\beta W \times 2\beta H$ region, where the values range from $0$ to $\Phi$, indicating the mixing ratio for low-frequency amplitude. $\Phi$ is a parameter that increases during model training: $\Phi(t)=\frac{t}{t_{total}}$, where $t$ denotes the current training iteration and $t_{total}$ is the maximum training step, controlling the gradual enhancement of low-frequency components of unlabeled data in the intermediate samples.

\begin{table*}[t]
\centering
\footnotesize
\caption{Comparison of different methods on Fundus dataset. \#L represents the number of labeled samples. In the line of Upper bound, * denotes the Upper bound using all training samples in a domain as labeled data. $\uparrow$ indicates that a higher value corresponds to better performance, while $\downarrow$ suggests the opposite. The best performance is marked as \textbf{bold}, and the second-best is \underline{underlined}.}
\begin{tabular}{ll|c|cccc|cccc}
    \hline
    \multicolumn{3}{c|}{Task} & \multicolumn{8}{c}{Optic Cup / Disc Segmentation}\\
    \hline
    \multicolumn{2}{c|}{\multirow{2}{*}{Method}} & \multirow{2}{*}{\#L} & \multicolumn{4}{c|}{DC $\uparrow$} & DC $\uparrow$ & JC $\uparrow$ & HD $\downarrow$ & ASD $\downarrow$\\
    \cline{4-11}
    & & & Domain 1 & Domain 2 & Domain 3 & Domain 4 & Avg. & Avg. & Avg. & Avg.\\
    \hline
    U-Net & & 20 & 59.54 / 73.89 & 71.28 / 74.23 & 50.87 / 64.29 & 35.61 / 63.30 & 61.63 & 52.65 & 48.28 & 28.86\\
    UA-MT & \pub{MICCAI'19} & 20 & 59.35 / 78.46 & 63.08 / 74.45 & 35.24 / 47.73 & 36.18 / 55.43 & 56.24 & 47.00 & 48.64 & 31.35\\
    FDA & \pub{CVPR'20} & 20 & 76.99 / 89.94 & \underline{77.69} / 89.63 & 78.27 / 90.96 & 64.52 / 74.29 & 80.29 & 71.05 & 16.23 & 8.44\\
    SIFA & \pub{TMI'20} & 20 &50.67 / 75.30 & 64.44 / 80.69 & 61.67 / 83.77 & 55.07 / 70.67 & 67.78 & 54.77 & 20.16 & 10.93\\
    FixMatch & \pub{NeurIPS'20} & 20 & \underline{81.18} / \underline{91.29} & 72.04 / 87.60 & \underline{80.41} / \underline{92.95} & 74.58 / 87.07 & \underline{83.39} & 73.48 & 11.77 & \underline{5.60}\\
    CPS & \pub{CVPR'21} & 20 & 64.53 / 86.25 & 70.26 / 86.97 & 42.92 / 54.94 & 36.98 / 46.70 & 61.19 & 52.69 & 34.44 & 26.79\\
    CoraNet & \pub{TMI'21} & 20 & 61.64 / 87.32 & 65.56 / 87.05 & 66.12 / 83.54 & 49.01 / 77.73 & 72.25 & 60.50 & 20.52 & 10.44\\
    UDA-VAE++ & \pub{CVPR'22} & 20 & 55.01 / 80.76 & 68.87 / 85.94 & 63.23 / 84.92 & 68.42 / 80.89 & 73.51 & 61.40 & 17.60 & 9.86\\
    SS-Net & \pub{MICCAI'22} & 20 & 59.42 / 78.15 & 67.32 / 85.05 & 45.69 / 69.91 & 38.76 / 61.13 & 63.18 & 53.49 & 44.90 & 25.73\\
    BCP & \pub{CVPR'23} & 20 & 71.65 / 91.10 & 77.19 / \textbf{92.00} & 72.63 / 90.77 & \underline{77.67} / \underline{91.42} & 83.05 & \underline{73.66} & \underline{11.05} & 5.80\\
    CauSSL & \pub{ICCV'23} & 20 & 63.38 / 80.60 & 67.52 / 80.72 & 49.53 / 63.88 & 39.43 / 49.43 & 61.81 & 51.80 & 41.25 & 23.94\\
    \hline
    Ours & \pub{this paper} & 20 & \textbf{83.71} / \textbf{92.96} & \textbf{80.47} / \underline{89.93} & \textbf{84.18} / \textbf{92.97} & \textbf{83.71} / \textbf{93.38} & \textbf{87.66} & \textbf{79.10} & \textbf{8.21} & \textbf{3.89}\\
    \color{gray} Upper bound & & \color{gray} * & \color{gray} 85.53 / \color{gray} 93.41 & \color{gray} 80.55 / \color{gray} 90.90 & \color{gray} 85.44 / \color{gray} 93.04 & \color{gray} 85.61 / \color{gray} 93.21 & \color{gray} 88.46 & \color{gray} 80.35 & \color{gray} 7.41 & \color{gray} 3.70\\
    \hline
\end{tabular}
\label{fundus}
\end{table*}

\begin{figure*}[!t]
\centering
\includegraphics[width=1.0\linewidth]{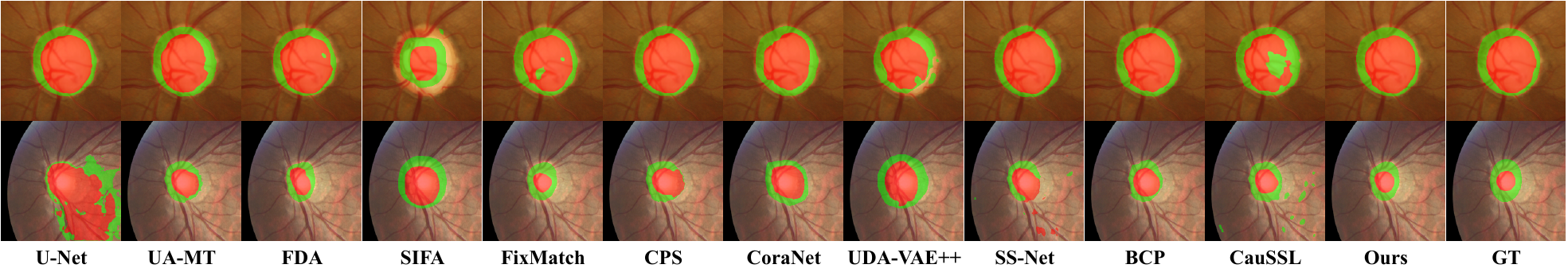}
\caption{Visual results from Fundus dataset. The first row shows results for a test sample from the labeled domain (Domain 1), and the second row displays results for one from another domain (Domain 4). Red and green represent the Optical Cup and Disc, respectively.} 
\label{fundus_img}
\vspace{-10pt}
\end{figure*}
\subsection{Loss Function}
For each pair of labeled and unlabeled data, the overall training objective function consists of two parts. The supervised loss is trivial, while the unsupervised loss is calculated through consistency regularization of symmetric guidance:
\begin{equation}
    \label{total_loss}
    \mathcal{L}_{total}=\mathcal{L}_s+\lambda(\mathcal{L}_{in}+\mathcal{L}_{out}+\lambda\mathcal{L}_{sym}),
\end{equation}
where $\lambda$ is a weight coefficient decided by a time-dependent Gaussian warming-up function: $\lambda(t)=e^{-5(1-t/t_{total})}$. Noting that the coefficient $\lambda^2$ for $\mathcal{L}_{sym}$ increases relatively slowly in the early stages of training because the quality of pseudo labels generated through mergence depends on the effective training of $\mathcal{L}_{in}$ and $\mathcal{L}_{out}$.

\subsection{Discussion of Inference and Parameter Size}
Following previous work~\cite{tarvainen2017mean}, our method consists of two models, both implemented based on U-Net~\cite{ronneberger2015u}, without introducing any additional network structures. We promote domain knowledge transfer through a training strategy and data augmentation to mitigate error accumulation in SSMS caused by domain shift. During the testing phase, given a test image $x^{test}$, we obtain the probability map $p^{test}=f^s(x^{test})$ through the student model and subsequently derive the segmentation result. The inference parameters only involve those from the student model, resulting in efficiency comparable to U-Net~\cite{ronneberger2015u}.

\section{Experiments}
\begin{table*}[h]
\centering
\footnotesize
\caption{Comparison of different methods on Prostate dataset.}
\begin{tabular}{ll|c|cccccc|cccc}
    \hline
    \multicolumn{3}{c|}{Task} & \multicolumn{10}{c}{Prostate Segmentation}\\
    \hline
    \multicolumn{2}{c|}{\multirow{2}{*}{Method}} & \multirow{2}{*}{\#L} & \multicolumn{6}{c|}{DC $\uparrow$} & DC $\uparrow$ & JC $\uparrow$ & HD $\downarrow$ & ASD $\downarrow$\\
    \cline{4-13}
    & & & RUNMC & BMC & HCRUDB & UCL & BIDMC & HK & Avg. & Avg. & Avg. & Avg.\\
    \hline
    U-Net & & 40 & 31.11 & 35.07 & 20.04 & 38.18 & 19.41 & 26.62 & 28.41 & 23.24 & 95.11 & 65.84\\
    UA-MT & \pub{MICCAI'19} & 40 & 29.44 & 4.68 & 12.49 & 39.42 & 17.94 & 18.22 & 20.37 & 14.88 & 112.07 & 77.58\\
    FDA & \pub{CVPR'20} & 40 & 47.44 & 35.37 & 24.54 & 61.01 & 28.19 & 40.51 & 39.51 & 32.17 & 76.67 & 47.87\\
    SIFA & \pub{TMI'20} & 40 & 72.67 & 70.37 & 64.08 & 73.49 & 71.62 & 65.16 & 69.57 & 56.78 & 29.43 & \underline{13.03}\\
    FixMatch & \pub{NeurIPS'20} & 40 & \underline{83.58} & 69.17 & \underline{73.63} & \underline{79.21} & 56.07 & \underline{84.78} & \underline{74.41} & \underline{65.96} & \underline{24.18} & 14.09\\
    CPS & \pub{CVPR'21} & 40 & 29.83 & 9.21 & 11.84 & 43.84 & 13.51 & 14.56 & 20.47 & 15.12 & 115.96 & 78.51\\
    CoraNet & \pub{TMI'21} & 40 & 69.43 & 31.16 & 16.29 & 69.33 & 24.66 & 22.16 & 38.84 & 31.48 & 67.91 & 44.98\\
    UDA-VAE++ & \pub{CVPR'22} & 40 & 68.73 & 69.36 & 65.49 & 67.19 & 63.29 & 65.15 & 66.54 & 52.80 & 34.20 & 15.48\\
    SS-Net & \pub{MICCAI'22} & 40 & 29.10 & 13.49 & 14.20 & 51.96 & 23.83 & 13.23 & 24.30 & 18.74 & 109.54 & 71.13\\
    BCP & \pub{CVPR'23} & 40 & 70.15 & \underline{71.97} & 46.15 & 58.93 & \underline{74.21} & 67.47 & 64.81 & 55.17 & 52.60 & 27.22\\
    CauSSL & \pub{ICCV'23} & 40 & 24.10 & 27.46 & 16.94 & 27.23 & 15.28 & 14.56 & 20.93 & 15.48 & 114.62 & 73.30\\
    \hline
    Ours & \pub{this pape} & 40 & \textbf{88.76} & \textbf{86.35} & \textbf{87.61} & \textbf{88.34} & \textbf{88.62} & \textbf{88.20} & \textbf{87.98} & \textbf{80.21} & \textbf{10.36} & \textbf{4.20}\\
    \color{gray} Upper bound & & \color{gray} * & \color{gray} 88.52 & \color{gray} 88.61 & \color{gray} 85.71 & \color{gray} 88.61 & \color{gray} 88.98 & \color{gray} 89.49 & \color{gray} 88.32 & \color{gray} 80.71 & \color{gray} 10.05 & \color{gray} 4.12\\
    \hline
\end{tabular}
\label{prostate}
\end{table*}

\subsection{Datasets}
\textbf{Fundus dataset}~\cite{wang2020dofe} comprises retinal fundus images collected from four medical centers, primarily for optic cup and disc segmentation tasks. Each image has been cropped to form a region of interest with a $800 \times 800$ bounding-box. We resize and randomly crop images to $256 \times 256$.

\textbf{Prostate dataset}~\cite{liu2020shape} contains prostate T2-weighted MRI data (with segmentation mask) collected from six different data sources out of three public datasets. We randomly split the dataset into training and testing sets based on a ratio of $4:1$, with each 2D slice resized and randomly cropped to $384 \times 384$. Labeled samples are selected from consecutive slices within cases, with at most one case of intersection and no slice intersection with unlabeled samples.

\textbf{M\&Ms dataset}~\cite{campello2021multi} contains 320 subjects scanned by four different magnetic resonance scanner vendors, primarily for the left ventricle (LV), left ventricle myocardium (MYO), and right ventricle (RV) segmentation tasks. Only the end-systole and end-diastole phases are annotated. We divide the data with annotation into training and testing sets following the same criteria as Prostate dataset. Each slice is resized and randomly cropped to $288 \times 288$.

There are $20$ labeled data in Fundus dataset and M\&Ms dataset, and $40$ labeled data in Prostate dataset. We normalized the data to [-1,1]. The detailed dataset partition information is presented in the supplementary material.

\begin{table*}[t]
\setlength\tabcolsep{1.0mm}
\centering
\footnotesize
\caption{Comparison of different methods on M\&Ms dataset.}
\begin{tabular}{ll|c|cccc|cccc}
    \hline
    \multicolumn{3}{c|}{Task} & \multicolumn{8}{c}{LV / MYO / RV Segmentation}\\
    \hline
    \multicolumn{2}{c|}{\multirow{2}{*}{Method}} & \multirow{2}{*}{\#L} & \multicolumn{4}{c|}{DC $\uparrow$} & DC $\uparrow$ & JC $\uparrow$ & HD $\downarrow$ & ASD $\downarrow$\\
    \cline{4-11}
    & & & Vendor A & Vendor B & Vendor C & Vendor D & Avg. & Avg. & Avg. & Avg.\\
    \hline
    U-Net & & 20 & 57.29 / 37.85 / 34.65 & 73.44 / 64.20 / 53.58 & 55.83 / 48.47 / 44.84 & 63.85 / 52.25 / 49.85 & 53.01 & 44.30 & 38.07 & 22.88\\
    UA-MT & \pub{MICCAI'19} & 20 & 38.02 / 25.51 / 14.94 & 61.85 / 54.27 / 47.33 & 43.13 / 35.66 / 28.54 & 41.89 / 38.25 / 26.11 & 37.96 & 29.14 & 72.35 & 40.84\\
    FDA & \pub{CVPR'20} & 20 & 61.66 / 36.32 / 34.71 & 80.67 / 70.99 / 56.75 & 73.80 / 63.62 / 58.36 & 77.23 / 68.87 / 64.33 & 62.28 & 53.33 & 25.99 & 16.10\\
    SIFA & \pub{TMI'20} & 20 & 63.97 / 34.64 / 37.25 & 66.46 / 44.97 / 43.22 & 56.52 / 39.63 / 38.55 & 59.57 / 40.63 / 40.26 & 47.14 & 34.96 & 25.01 & 11.45\\
    FixMatch & \pub{NeurIPS'20} & 20 & \underline{87.26} / \textbf{77.78} / 77.14 & \underline{91.06} / \underline{82.78} / \underline{79.07} & \underline{87.84} / \underline{80.07} / \underline{78.03} & \underline{90.86} / \underline{81.75} / \underline{81.84} & \underline{82.96} & \underline{73.99} & \underline{6.21} & \underline{3.51}\\
    CPS & \pub{CVPR'21} & 20 & 46.40 / 29.01 / 16.70 & 71.48 / 63.08 / 49.39 & 44.38 / 39.43 / 32.42 & 47.71 / 40.75 / 29.75 & 42.54 & 33.82 & 58.30 & 34.94\\
    CoraNet & \pub{TMI'21} & 20 & 65.70 / 27.79 / 22.16 & 63.32 / 48.63 / 46.56 & 64.89 / 48.59 / 45.30 & 68.38 / 55.88 / 46.79 & 50.33 & 40.54 & 32.98 & 19.22\\
    UDA-VAE++ & \pub{CVPR'22} & 20 & 51.14 / 36.20 / 12.99 & 71.95 / 53.16 / 36.68 & 57.88 / 41.64 / 30.19 & 31.71 / 27.32 / 20.48 & 39.28 & 28.82 & 53.90 & 24.94\\
    SS-Net & \pub{MICCAI'22} & 20 & 48.98 / 27.02 / 20.65 & 59.67 / 52.69 / 40.86 & 54.92 / 41.40 / 39.62 & 55.04 / 52.07 / 35.88 & 44.07 & 35.89 & 49.50 & 32.55\\
    BCP & \pub{CVPR'23} & 20 & 85.91 / 73.82 / \underline{78.08} & 85.66 / 74.85 / 76.04 & 61.61 / 54.05 / 51.87 & 76.57 / 62.22 / 79.16 & 71.65 & 62.67 & 30.91 & 18.22\\
    CauSSL & \pub{ICCV'23} & 20 & 40.20 / 21.93 / 10.46 & 50.99 / 42.66 / 31.94 & 41.05 / 34.00 / 29.95 & 53.78 / 37.92 / 30.44 & 35.44 & 26.73 & 72.90 & 37.99\\
    \hline
    Ours & \pub{this paper} & 20 & \textbf{87.77} / \underline{76.36} / \textbf{80.65} & \textbf{91.48} / \textbf{83.68} / \textbf{81.46} & \textbf{89.25} / \textbf{82.65} / \textbf{82.27} & \textbf{90.91} / \textbf{82.34} / \textbf{82.86} & \textbf{84.31} & \textbf{75.18} & \textbf{5.15} & \textbf{2.42}\\
    \color{gray} Upper bound & & \color{gray} * & \color{gray} 91.37 / \color{gray} 83.61  / \color{gray} 84.13 & \color{gray} 92.33 / \color{gray} 85.12  / \color{gray} 84.43 & \color{gray} 92.20 / \color{gray} 84.43  / \color{gray} 84.56 & \color{gray} 92.08 / \color{gray} 83.78  / \color{gray} 83.77 & \color{gray} 86.82 & \color{gray} 78.64 & \color{gray} 4.32 & \color{gray} 2.14\\
    \hline
\end{tabular}
\label{mnms}
\vspace{-5pt}
\end{table*}


\begin{table*}[t]
\centering
\footnotesize
\caption{Ablation experiments on Fundus dataset.}
\begin{tabular}{c|ccccc|ccccc}
    \hline
    \multicolumn{6}{c|}{Task} & \multicolumn{5}{c}{Optic Cup / Disc Segmentation}\\
    \hline
    \multirow{2}{*}{Method} & \multirow{2}{*}{UCP} & \multirow{2}{*}{VanillaGD} & \multirow{2}{*}{SymGD} & \multirow{2}{*}{TP-RAM} & \multirow{2}{*}{RAM} & \multicolumn{5}{c}{DC $\uparrow$}\\
    \cline{7-11}
    & & & & & & Domain 1 & Domain 2 & Domain 3 & Domain 4 & Avg.\\
    \hline
    \#1 & \checkmark & & & & & 82.67 / 93.11 & 72.08 / 88.86 & 82.97 / 92.78 & 80.84 / 92.94 & 85.78\\
    \#2 & \checkmark & \checkmark & & & & 83.19 / 93.45 & 73.57 / 89.48 & 82.16 / 92.91 & 80.42 / 93.35 & 86.07\\
    \#3 & \checkmark & & \checkmark & & & 83.21 / 93.48 & 76.13 / 89.06 & 83.04 / 92.87 & 83.63 / \textbf{93.69} & 86.89\\
    \#4 & \checkmark & & & \checkmark & & 83.41 / \textbf{93.54} & 77.18 / 88.96 & 82.69 / 92.88 & 83.54 / 93.39 & 86.95\\
    \#5 & \checkmark & & \checkmark & & \checkmark & 83.27 / 93.41 & 76.14 / 88.46 & 83.27 / 92.90 & 83.64 / 93.40 & 86.82\\
    \hline
    \#6 & \checkmark & & \checkmark & \checkmark & & \textbf{83.71} / 92.96 & \textbf{80.47} / \textbf{89.93} & \textbf{84.18} / \textbf{92.97} & \textbf{83.71} / 93.38 & \textbf{87.66}\\
    \hline
\end{tabular}
\label{fundusablation}
\end{table*}
\begin{table}[t]
\centering
\small
\caption{Varying the low-frequency region size parameter $\beta$ on Fundus dataset.}
\begin{tabular}{c|cccc}
    \hline
    $\beta$ & 0.1 & 0.05 & 0.01 & 0.005\\
    \hline
    Avg. DC $\uparrow$ & 87.32 & 87.49 & \textbf{87.66} & 87.52\\
    \hline
\end{tabular}
\label{beta}
\vspace{-10pt}
\end{table}


\subsection{Implementation Details}
Our method is implemented based on Pytorch and an NVIDIA GeForce RTX 4090 GPU. We set $\beta=0.01$, $\tau=0.95$ as the default value in experiments. We use Stochastic Gradient Descent (SGD) with a momentum of 0.9 and weight decay of 0.0001 as optimizer, with an initial learning rate of 0.03. The batch size is set as 8, including 4 labeled data and 4 unlabeled data. The iteration is set to 30,000 for Fundus dataset and 60,000 for Postate and M\&Ms dataset. In the testing stage, the final segmentation results are determined by the student model. We compare our method with other state-of-the-art (SOTA) methods, including supervised methods such as UA-MT~\cite{yu2019uncertainty}, FixMatch~\cite{sohn2020fixmatch}, CPS~\cite{chen2021semi}, CoraNet~\cite{shi2021inconsistency}, SS-Net~\cite{wu2022exploring}, BCP~\cite{bai2023bidirectional}, and CauSSL~\cite{miao2023caussl} as well as domain unsupervised adaptation methods like FDA~\cite{yang2020fda}, SIFA~\cite{chen2020unsupervised} and UDA-VAE++~\cite{lu2022unsupervised}. 
In each experiment, a small amount of data from one domain is labeled (\eg, Domain 1 in \cref{fundus}), while the remainder serves as unlabeled data.
For the upper bound, we employ UCP in FixMatch, and use all available training data from a certain domain as labeled data, providing the model with sufficient source domain information. 
We adopt the Dice coefficient (DC), Jaccard coefficient (JC), 95\% Hausdorff Distance (HD), and Average Surface Distance (ASD) as evaluation metrics. Except for SIFA, which uses ResNet blocks~\cite{he2016deep} for the generator and decoder, all other methods use U-Net~\cite{ronneberger2015u} as the backbone. 

\subsection{Comparison with State-of-the-Art Methods}
\textbf{Results on Fundus dataset.} We conduct experiments on Fundus dataset with only 20 labeled data, which is a dual object segmentation task with overlapping regions. As shown in \cref{fundus}, our method consistently outperforms other approaches in all metrics, and exhibits a minimal gap from the upper bound. Some SSMS methods exhibit performance degradation due to error accumulation, to the extent that their performance is even worse than training with labeled data alone. UDA methods face challenges in achieving effective knowledge transfer with limited labeling and multiple target domains. The segmentation result examples are shown in~\cref{fundus_img}, demonstrating that our method achieves superior segmentation for samples from the same domain as labeled data and those from different domains.

\textbf{Results on Prostate dataset.} In the case of Prostate dataset, which is a single object segmentation task, our method demonstrates a more significant advantage and increased robustness when trained with 40 labeled data. The disparity from the upper bound is also minimal. Given the larger differences between domains in Prostate dataset, SSMS methods are notably influenced by more severe error accumulation. UDA methods also demonstrate poor performance in this scenario. The segmentation result examples are presented in the supplementary material.

\textbf{Results on M\&Ms dataset.} We conducted experiments on M\&Ms dataset using 20 labeled data. The results in \cref{mnms} demonstrate that our approach is effective in multi-object segmentation tasks, achieving optimal performance. Due to the abundance of samples in M\&Ms dataset, the substantial increase in labeled data significantly enhances the performance of the upper bound. The segmentation result examples are presented in the supplementary material.

It is worth noting that while BCP considers the distribution difference between labeled and unlabeled data in SSMS, our method still exhibits significant advantages in the scenario where there is distribution difference within unlabeled data and between unlabeled and labeled data. We also compare our method with more SSMS methods aimed at mitigating distribution differences, and detailed analysis results can be found in the supplementary material.

\subsection{Ablation Study}

In this paragraph, to validate the effectiveness of each component, we conduct a series of ablation experiments on Fundus dataset. The investigated settings are introduced as follows: 1) UCP: Generating intermediate samples based on UCP, 2) VanillaGD: Guiding $p^s$ with $\hat{p}$, 3) SymGD: Guiding $p^s$ with mergence of $\hat{p}_{out}$ and $\hat{p}_{in}$, 4) TP-RAM: Gradually increasing the range of value for $\beta$, 5) RAM: $\beta$ is uniformly sampled from the distribution in the range from $0$ to $1$.
The results are shown in \cref{fundusablation}.

\textbf{The Efficiency of SymGD.} Compared to \#1, \#3 introduces an additional guidance branch that guides the predictions of unlabeled data by integrating pseudo labels from dual perspectives. SymGD effectively leverages the intermediate domain information, aiding the model in making more accurate predictions for unlabeled data. In \#2, we use the pseudo label $\hat{p}$ generated from the teacher model to guide the student model's prediction $p^s$, and this straightforward additional consistency constraint has a limited impact on the model performance improvement.

\textbf{The Efficiency of TP-RAM.} Moreover, in \#4, we gradually introduce stylistic transition components to intermediate samples, not limited to local semantic fusion. The comprehensive and stable construction of intermediate domains facilitates the transfer of domain knowledge from the domain of labeled data to the domains of unlabeled data. In \#5, we replace the TP-RAM in our method with RAM, introducing stylistic transition arbitrarily for intermediate samples. Training with significant stylistic differences from labeled data in the early stages is detrimental to the stable construction of the intermediate domains, and compared to \#2, it might even harm the model's performance.

In \#6, we incorporate two new modules to form a unified domain knowledge transfer framework, ultimately achieving the best performance in the MiDSS scenarios.

\textbf{Size of Low-frequency Region in $M_\beta$.} We also explore the relationship between the selection of low-frequency region size and model performance. Here, $M_\beta$ represents the low-frequency region of an image with a size of $2\beta W \times 2\beta H$, $\beta\in\{0.1,0.005,0.01,0.005\}$. \cref{beta} reveals that the model is quite robust to the choice of $\beta$. Overall, there is a slight trend of performance improvement followed by a decline, with the best performance achieved when $\beta=0.01$.

\section{Conclusion}
In this paper, we explore a novel and practical setting, MiDSS, which involves the simultaneous challenges of limited labeled data and domain shift.
To address the issue of declining pseudo label quality due to domain shift, we construct intermediate domains by UCP. Additionally, we introduce SymGD to enhance the utilization of intermediate domains information. Considering the stylistic differences between different domains, we design TP-RAM to introduce comprehensive and stable style transition components to intermediate domains. Extensive results on three datasets demonstrate the effectiveness of our method.


{
    \small
    \bibliographystyle{ieeenat_fullname}
    \bibliography{main}
}


\end{document}


\clearpage
\setcounter{page}{1}
\title{Constructing and Exploring Intermediate Domains in Mixed Domain Semi-supervised Medical Image Segmentation}
\maketitlesupplementary

\renewcommand{\thesection}{\Alph{section}}
\setcounter{section}{0}

\begin{table}[!h]
  \centering
    \footnotesize
  \begin{tabular}{l|cc|c}
    \toprule
    \multirow{2}{*}{Dataset} & \multicolumn{2}{c|}{Training} & \multirow{2}{*}{\#Testing} \\
    \cline{2-3}
    & \#Labeled & \#Unlabeled &\\
    \midrule
    Fundus & 20 & 769 & 271 \\
    Prostate & 40 & 1,470 & 357 \\
    M\&Ms & 20 & 3,427 & 863\\
    \bottomrule
  \end{tabular}
  \caption{Detailed partition information of three datasets. For each dataset, \#Labeled, \#Unlabeled, and \#Testing indicate the number of labeled data, unlabeled data and test data, respectively.}
  \label{partition}
\end{table}
\begin{table*}[hb]
\centering
\footnotesize
\caption{Comparison of different methods on Fundus dataset. + denotes we employ this method based on FixMatch. \#L represents the number of labeled samples. $\uparrow$ indicates that a higher value corresponds to better performance, while $\downarrow$ suggests the opposite. The best performance is marked as \textbf{bold}, and the second-best is \underline{underlined}.}
\begin{tabular}{l|c|cccc|cccc}
    \hline
    \multicolumn{2}{c|}{Task} & \multicolumn{8}{c}{Optic Cup / Disc Segmentation}\\
    \hline
    \multirow{2}{*}{Method} & \multirow{2}{*}{\#L} & \multicolumn{4}{c|}{DC $\uparrow$} & DC $\uparrow$ & JC $\uparrow$ & HD $\downarrow$ & ASD $\downarrow$\\
    \cline{3-10}
    & & Domain 1 & Domain 2 & Domain 3 & Domain 4 & Avg. & Avg. & Avg. & Avg.\\
    \hline
    FixMatch & 20 & 81.18 / 91.29 & 72.04 / 87.60 & 80.41 / \underline{92.95} & 74.58 / 87.07 & 83.39 & 73.48 & 11.77 & 5.60\\
    +FDA & 20 & 82.59 / 92.80 & 74.34 / 88.63 & 80.08 / 92.64 & 77.66 / 88.99 & 84.72 & 75.33 & 10.38 & 4.82\\
    +CutMix & 20 & \underline{83.62} / 92.75 & 71.45 / 88.69 & 82.09 / 92.23 & 80.57 / 93.30 & 85.59 & 76.32 & 9.61 & 4.71\\
    +ClassMix & 20 & 71.35 / 89.47 & \underline{76.25} / \underline{89.54} & 83.01 / 90.95 & 81.41 / 92.81 & 84.35 & 75.02 & 10.84 & 5.59\\
    +CowMix & 20 & 83.54 / 92.72 & 71.76 / 88.42 & \underline{83.15} / 92.13 & \underline{83.05} / 93.13 & \underline{85.99} & \underline{77.07} & 9.28 & 4.56\\
    +FMix & 20 & 81.88 / \underline{92.90} & 72.96 / 89.10 & 82.41 / 92.80 & 82.19 / \underline{93.33} & 85.95 & 76.80 & \underline{9.26} & \underline{4.52}\\
    \hline
    Ours & 20 & \textbf{83.71} / \textbf{92.96} & \textbf{80.47} / \textbf{89.93} & \textbf{84.18} / \textbf{92.97} & \textbf{83.71} / \textbf{93.38} & \textbf{87.66} & \textbf{79.10} & \textbf{8.21} & \textbf{3.89}\\
    \hline
\end{tabular}
\label{fundus_ssmsuda}
\end{table*}

\begin{table*}[hb]
\centering
\footnotesize
\caption{Comparison of different methods on Prostate dataset.}
\begin{tabular}{l|c|cccccc|cccc}
    \hline
    \multicolumn{2}{c|}{Task} & \multicolumn{10}{c}{Prostate Segmentation}\\
    \hline
    \multirow{2}{*}{Method} & \multirow{2}{*}{\#L} & \multicolumn{6}{c|}{DC $\uparrow$} & DC $\uparrow$ & JC $\uparrow$ & HD $\downarrow$ & ASD $\downarrow$\\
    \cline{3-12}
    & & RUNMC & BMC & HCRUDB & UCL & BIDMC & HK & Avg. & Avg. & Avg. & Avg.\\
    \hline
    FixMatch & 40 & 83.58 & 69.17 & 73.63 & 79.21 & 56.07 & 84.78 & 74.41 & 65.96 & 24.18 & 14.09\\
    +FDA & 40 & 77.78 & 80.89 & 57.47 & 85.07 & 33.31 & 78.96 & 68.91 & 63.13 & 40.35 & 21.77\\
    +CutMix & 40 & 86.97 & \underline{85.23} & 81.63 & 87.26 & 87.62 & \underline{85.39} & 85.68 & \underline{78.10} & 12.77 & 5.94\\
    +ClassMix & 40 & 85.02 & 69.16 & 69.06 & 85.32 & 43.16 & 76.03 & 71.29 & 60.70 & 57.52 & 28.24\\
    +CowMix & 40 & 86.45 & 85.05 & 83.68 & \underline{87.75} & \underline{88.20} & 84.41 & \underline{85.92} & 78.03 & \underline{12.56} & \underline{5.32}\\
    +FMix & 40 & \underline{87.59} & 84.80 & \underline{84.95} & 87.10 & 88.15 & 75.48 & 84.19 & 76.37 &  14.54& 6.55\\
    \hline
    Ours & 40 & \textbf{88.76} & \textbf{86.35} & \textbf{87.61} & \textbf{88.34} & \textbf{88.62} & \textbf{88.20} & \textbf{87.98} & \textbf{80.21} & \textbf{10.36} & \textbf{4.20}\\
    \hline
\end{tabular}
\label{prostate_ssmsuda}
\end{table*}

\begin{table*}[thb]
\setlength\tabcolsep{1.0mm}
\centering
\footnotesize
\caption{Comparison of different methods on M\&Ms dataset.}
\begin{tabular}{l|c|cccc|cccc}
    \hline
    \multicolumn{2}{c|}{Task} & \multicolumn{8}{c}{LV / MYO / RV Segmentation}\\
    \hline
    \multirow{2}{*}{Method} & \multirow{2}{*}{\#L} & \multicolumn{4}{c|}{DC $\uparrow$} & DC $\uparrow$ & JC $\uparrow$ & HD $\downarrow$ & ASD $\downarrow$\\
    \cline{3-10}
    & & Vendor A & Vendor B & Vendor C & Vendor D & Avg. & Avg. & Avg. & Avg.\\
    \hline
    FixMatch & 20 & \underline{87.26} / \underline{77.78} / 77.14 & 91.06 / 82.78 / 79.07 & 87.84 / 80.07 / 78.03 & \underline{90.86} / \underline{81.75} / \underline{81.84} & 82.96 & 73.99 & 6.21 & 3.51\\
    +FDA & 20 & 85.22 / 75.40 / 76.30 & 89.91 / 81.59 / 78.93 & 85.26 / 77.32 / 74.44 & 89.74 / 81.60 / 80.20 & 81.33 & 72.12 & 7.09 & 4.07\\
    +CutMix & 20 & 86.87 / 76.90 / \underline{80.01} & 91.12 / 82.04 / 79.94 & 87.65 / 81.31 / \underline{80.42} & 90.06 / 81.73 / 81.60 & \underline{83.30} & \underline{74.45} & \underline{5.53} & \underline{2.87}\\
    +ClassMix & 20 & 65.81 / 66.18 / 73.98 & 89.84 / 81.48 / \underline{80.94} & \underline{88.22} / \underline{81.96} / 80.00 & 85.87 / 79.02 / 80.51 & 79.48 & 70.02 & 16.98 & 8.41\\
    +CowMix & 20 & 87.16 / \textbf{78.25} / 78.46 & 91.10 / 82.65 / 77.98 & 87.43 / 80.45 / 79.20 & 90.38 / 81.28 / 80.71 & 82.92 & 73.80 & 6.37 & 3.48\\
    +FMix & 20 & 86.44 / 75.16 / 79.42 & \underline{91.20} / \underline{82.84} / 79.34 & 87.65 / 81.05 / 80.39 & 90.39 / 81.72 / 81.57 & 83.10 & 73.87 & 5.78 & 2.93\\
    \hline
    Ours & 20 & \textbf{87.77} / 76.36 / \textbf{80.65} & \textbf{91.48} / \textbf{83.68} / \textbf{81.46} & \textbf{89.25} / \textbf{82.65} / \textbf{82.27} & \textbf{90.91} / \textbf{82.34} / \textbf{82.86} & \textbf{84.31} & \textbf{75.18} & \textbf{5.15} & \textbf{2.42}\\
    \hline
\end{tabular}
\label{mnms_ssmsuda}
\vspace{-5pt}
\end{table*}

\begin{figure*}[h]
\centering
\includegraphics[width=1.0\linewidth]{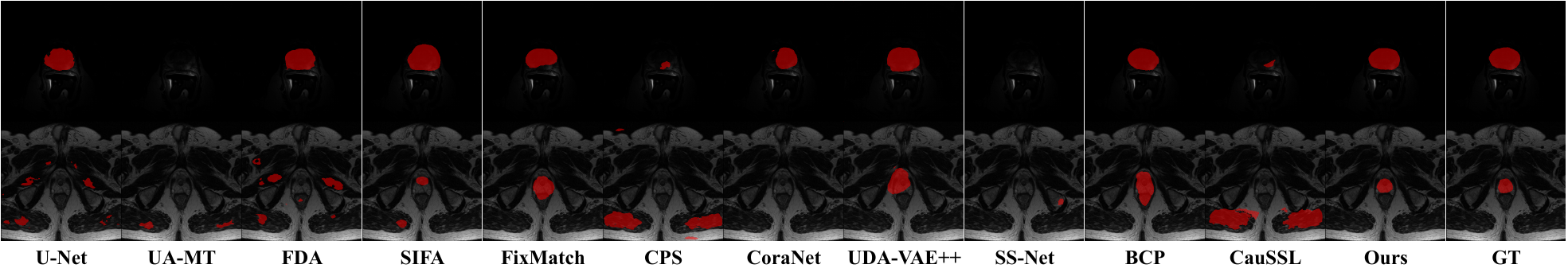}
\caption{Visual results from Prostate dataset.}
\label{prostate_img}
\end{figure*}

\begin{figure*}[th!]
\centering
\includegraphics[width=1.0\linewidth]{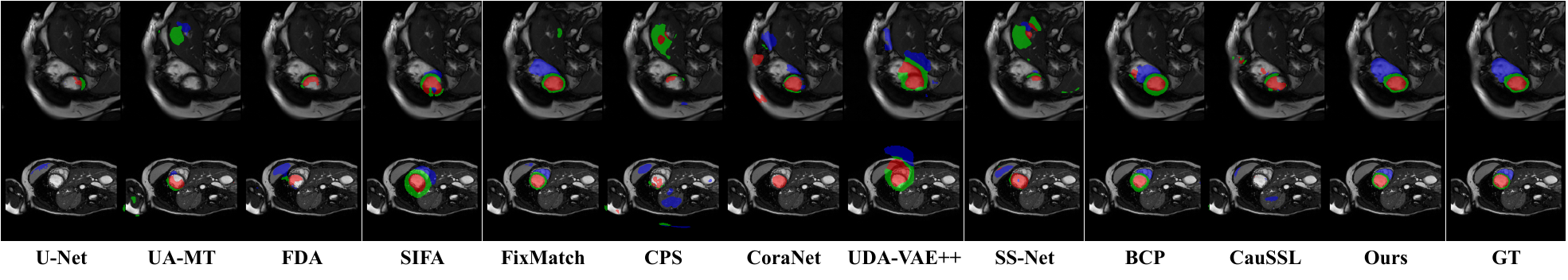}
\caption{Visual results from M\&Ms dataset. Red, green and blue represent LV, MYO and RV, respectively.}
\label{MNMS_img}
\end{figure*}

\vspace{-15pt}
\section{Detailed Dataset Partition}
The detailed description of the datasets is shown in \cref{partition}. In our setting, labeled data share a same distribution, while unlabeled data and testing set data come from multiple domains. Fundus dataset is inherently partitioned into training and testing sets. As for Prostate and M\&Ms datasets, we employed a 4:1 ratio for the division.

\section{Visual Results of Prostate and M\&Ms}
Being consistent with Fundus dataset, we present visual results under different methods for Prostate and M\&Ms datasets, as illustrated in \cref{prostate_img} and \cref{MNMS_img}, respectively. Due to error accumulation caused by distribution differences, many existing state-of-the-art methods exhibit inferior segmentation capabilities on test data with the same distribution as labeled data. Their performance degrades even further when tested on data from other domains. In contrast, our method demonstrates superior segmentation performance on test data from both the same and different domains as labeled data.

\section{Comparison with Methods Integrating Semi-supervised Medical Image Segmentation and Domain Adaptation}

Semi-supervised medical image segmentation (SSMS) methods and domain adaptation (DA) methods address distinct challenges in the Mixed Domain Semi-supervised Medical Image Segmentation scenario. For a fair comparison, we integrate various DA methods with the SSMS approach and evaluate their performance. Utilizing FixMatch~\cite{sohn2020fixmatch} as a baseline, we select FDA~\cite{yang2020fda}, CutMix~\cite{yun2019cutmix}, ClassMix~\cite{olsson2021classmix}, CowMix~\cite{french2020milking}, and FMix~\cite{harris2020fmix} to facilitate domain knowledge transfer. Specifically, FDA involves style transfer from labeled to unlabeled data, while other methods blend images using masks of different shapes. The results on three datasets are presented in \cref{fundus_ssmsuda,prostate_ssmsuda,mnms_ssmsuda}. In experiments on the Prostate dataset, we observed a significant performance drop when combining FDA and ClassMix with FixMatch. This emphasizes the necessity of thoughtfully selecting and combining of DA strategies to address the challenges posed by domain shift in SSMS. Additionally, the combination of CutMix and FMix with FixMatch consistently achieves superior performance on all three datasets. While constructing intermediate domains through local semantic mixing helps mitigate the adverse effects of the domain gap, the intermediate domains information has not been fully utilized. Moreover, it is crucial to note that a comprehensive intermediate domains construction should not be confined solely to mixing local semantics. Taking these observations into account, our method outperforms other methods on all three datasets.

{
    \small
    \bibliographystyle{ieeenat_fullname}
    \bibliography{main}
}